\pdfoutput=1

\documentclass[11pt]{article}

\usepackage[]{acl}
\usepackage{times}
\usepackage{latexsym}
\usepackage{amsmath}
\usepackage[T1]{fontenc}
\usepackage[utf8]{inputenc}
\usepackage{microtype}
\usepackage{inconsolata}
\usepackage{graphicx}
\usepackage{float}
\usepackage{subcaption}
\usepackage{booktabs}
\usepackage{enumitem}
\usepackage{cleveref}

\def\footurl#1{\footnote{\url{#1}}}

\title{Temporal Simultaneity Predicts Annotation Quality in Sentiment Corpora
}

\author{
  Idris Abdulmumin$^{1}$,
  Mokgadi Penelope Matloga$^{1}$,
  Tadesse Destaw Belay$^2$\\
  \bf Botshelo Kondowe$^{3}$,
  Letlhogonolo Mohleleng$^{3}$,
  Hareaipha Nkopo Letsoalo$^{3}$,\\
  \bf Shamsuddeen Hassan Muhammad$^{4}$, Vukosi Marivate$^{1}$\\[2mm]
  \footnotesize $^{1}$Data Science for Social Impact, University of Pretoria, $^{2}$Instituto Politécnico Nacional,\\
  \footnotesize $^{3}$Department of African Languages, University of Pretoria, $^{4}$Imperial College London
  \\[1mm]
  \footnotesize \texttt{Contact: idris.abdulmumin@up.ac.za}
}

\begin{document}
\maketitle

\begin{abstract}
  Annotation quality is difficult to sustain when campaigns span weeks or months with small annotator pools. We present a Setswana sentiment dataset of 3,565 tweets annotated by three native-speaker annotators across eight batches and examine why inter-annotator agreement (IAA) declines over time. Despite an aggregate Randolph's free-marginal Kappa of $\kappa = 0.76$, "excellent," per batch $\kappa$ falls by more than 32 points across the annotation task. Through six targeted analyses, we find that (i) label confusion concentrates on the negative/neutral boundary, (ii) two annotators show run-length drift consistent with \emph{autopilot} labeling, and (iii) the dominant predictor of $\kappa$ is \emph{temporal simultaneity}: tweets labeled within one minute achieve $\kappa = 0.98$, while those labeled more than a day apart reach only $\kappa = 0.65$. Annotation speed and tweet-level linguistic features show no meaningful association with $\kappa$. We benchmark three open multilingual encoders and proprietary models (GPT-5 and Gemini) on three-class sentiment classification; fine-tuning yields gains of 29 to 43 macro-F1 points over pretrained baselines, with GPT-5 few-shot leading overall (62.2 macro-F1). We release the dataset, per-annotation timestamps, and analysis code to support reproducible quality auditing for future African language NLP resources.
\end{abstract}

\section{Introduction}
\label{sec:intro}

The rapid growth of African-language NLP has produced a wave of labeled resources covering tasks from named-entity recognition to sentiment and emotion analysis \cite{muhammad-etal-2023-afrisenti,adelani-etal-2022-masakhaner}.
These datasets are often created with small volunteer or student annotator pools working over extended periods. These conditions are known to degrade annotation quality in high-resource settings  \citep{snow-etal-2008-cheap,sogaard-etal-2014-selection} and have rarely been studied for low-resource African languages.

A central challenge in low-resource dataset development is the trade-off between scale and quality. A small, carefully annotated corpus is generally more useful for training and evaluating downstream models than a larger, noisier one \cite{northcutt2021confident,aroyo2015truth}. Understanding the causes of quality erosion is therefore not a matter of methodological interest alone; it should directly inform annotation campaign design and the resulting dataset curation.


We present a new Setswana sentiment dataset and a systematic analysis of quality decline during the annotation campaign. The dataset comprises 3,565 tweets annotated by three native speakers across seven batches. Our annotation logs record the precise timestamp and other metadata for every annotation event, enabling analyses that go beyond the standard practice of reporting aggregate inter-annotator agreement (IAA). We investigate six complementary hypotheses about what causes inter-annotator disagreement:
\begin{enumerate}[label=\textbf{RQ\arabic*},leftmargin=*,noitemsep]
  \item Does IAA decline monotonically across batches?
  \item Which label pairs are most frequently confused, and does
    the confusion pattern shift over time?
  \item As the campaign progresses, do annotators exhibit \emph{autopilot} behavior: increasingly long runs of the same label?
  \item Is faster annotation (shorter inter-annotation gap) associated with
    lower IAA at the tweet or batch level?
  \item Does the time elapsed between the first and last annotator's timestamp for the same tweet (\textbf{\emph{temporal simultaneity}}) predict IAA?
  \item Do tweet-level linguistic features (e.g., token count) explain variation in IAA?
\end{enumerate}

\paragraph{Contributions}
\begin{itemize}[noitemsep]
  \item A manually annotated Setswana Twitter sentiment dataset of 3,565 tweets with full per-annotation timestamps (released publicly).
  \item A characterization of $\kappa$ decline over a multi-batch campaign, identifying temporal simultaneity as the dominant structural predictor.
  \item Null results for annotation speed and tweet-level linguistic complexity as predictors of $\kappa$, narrowing the search space for future mitigation strategies.
  \item Benchmarks on the dataset using various encoder and decoder models with concrete recommendations for annotation campaign design that require no additional budget.
\end{itemize}

\section{Related Work}
\label{sec:related}

\subsection{African-language sentiment and classification resources}
\citet{muhammad-etal-2023-afrisenti} introduced AfriSenti, a multilingual Twitter sentiment benchmark covering 14 African languages, providing baselines for few-shot and fine-tuned models. A shared task based on this data was later organized at SemEval-2023 \cite{muhammad-etal-2023-semeval}.
Earlier efforts include NaijaSenti \cite{muhammad-etal-2022-naijasenti}, covering Hausa, Igbo, Nigerian Pidgin, and Yor\`ub\'a; the Yor\`ub\'a Sentiment Corpus for movie reviews \cite{shode-etal-2022-yosm}; and SAfriSenti \cite{mabokela-schlippe-2022-sentiment}, a South African sentiment corpus covering Setswana and related languages. \citet{myoya-etal-2024-transport} analyzed commuter sentiment across multilingual public transport contexts in Kenya, Tanzania, and South Africa.

Beyond sentence- or document-level polarity, recent work has extended sentiment to aspect-level and dimensional representations. \citet{lee-etal-2026-dimabsa} released DimABSA, a multilingual multidomain dataset for aspect-based sentiment analysis that replaces categorical labels with real-valued affective dimensions, enabling more nuanced opinion modeling. \citet{becker-etal-2026-dimstance} similarly introduced DimStance, which models stance along dimensional valence and arousal axes across multiple languages.

More recent work has expanded from coarse sentiment polarity to include multi-label emotion classification. \citet{muhammad-etal-2025-brighter} released BRIGHTER, a human-annotated emotion recognition collection spanning 28 languages with emphasis on under-represented African languages;
\citet{belay-etal-2025-evaluating} introduced Ethio-Emo, a multi-label emotion dataset for four Ethiopian languages; \citet{adelani-etal-2023-masakhanews} released MasakhaNEWS, a topic classification benchmark across 16 African languages.

Across these efforts, IAA is typically reported as a single aggregate figure, without analysis of annotation quality trends or behavioral patterns.

\subsection{Annotation quality and IAA}
Inter-annotator agreement is most commonly measured by Cohen's \cite{cohen1960coefficient} or Fleiss' \cite{Fleiss1971-ay} $\kappa$, or Krippendorff's $\alpha$ \cite{krippendorff2011computing}.
For equal-frequency label distributions, \citet{randolph-kappa} proposed a free-marginal multi-rater Kappa that does not penalize annotators for collectively preferring one category, a property important for naturally skewed corpora such as sentiment datasets. We adopt Randolph's $\kappa$ throughout.
By convention \cite{randolph-kappa-calculator}, $\kappa < 0.40$ is "poor," $0.40 \le \kappa < 0.75$ is "intermediate to good," and $\kappa \ge 0.75$ is "excellent."
\citet{klie-etal-2024-analyzing} analyzes annotation quality practices in over 100 NLP dataset papers, and it is found that quality assurance is routinely under-reported.

\subsection{Annotation fatigue and drift}
\citet{snow-etal-2008-cheap} and subsequent crowdsourcing research \cite{kittur2008crowdsourcing,kazai2012search} document that annotator performance degrades with session length and repetition.
\citet{geva-etal-2019-modeling} show that NLU datasets harbor annotator-specific biases that accumulate over time.
In their work, \citet{muhammad-etal-2022-naijasenti} reported a visible decline in inter-annotator agreement across batches for Hausa, Igbo, and Yor\`ub\'a in the NaijaSenti corpus but did not investigate the causes of the decline.
In the African-language context, such longitudinal analyzes are almost entirely absent, which is the primary motivation for the approach we take here.

\subsection{Disagreement as signal}
A growing body of work argues that annotator disagreement should not be reflexively treated as noise to be resolved but as a meaningful signal encoding genuine label ambiguity, subjective interpretation, and demographic perspective \cite{fleisig-etal-2024-perspectivist,plank-2022-problem,uma-etal-2021-semeval,xu-jurgens-2026-beyond}.
We subscribe to this view, while also recognizing that not all disagreement is alike: some reflects genuine ambiguity in the data, while some reflects annotation process artifacts such as fatigue or scheduling. Distinguishing between the two is a goal in this work.

\section{The Setswana Sentiment Dataset}
\label{sec:dataset}

\subsection{Setswana Language}
\label{sec:language}

Setswana (ISO 639-3: \texttt{tsn}) is a Bantu language spoken by approximately 8 million people, predominantly in South Africa and Botswana, where it both holds official status. Based on the South African National Census of 2022, it is the fifth most-spoken home language in South Africa \cite{statsSA2025cultural}.
Like other Bantu languages, Setswana is agglutinative with extensive noun-class morphology, making tokenization and lexical-semantic analysis non-trivial.
Despite its speaker base, Setswana remains severely under-resourced in NLP: pre-trained language models and labeled corpora are scarce, and the language is absent from most mainstream benchmarks.

\subsection{Data Collection}
\label{sec:collection}

\paragraph{Collection method}
Tweets were collected from the public Twitter~API~(v2) using keyword and location filters targeting Setswana content between 2021 and 2022. A broad query was applied to maximize recall, with subsequent filtering applied.

\paragraph{Language identification}
All collected tweets were passed through the AfroLID language identification model \cite{adebara-etal-2022-afrolid}, which assigns confidence scores to the top-3 candidate languages.
We retained tweets for which Setswana (\texttt{tsn}) appeared among the top-2 predictions.
In the final annotated sample, 70.3\% of tweets have Setswana as the top-1 prediction (mean LID score 0.831) and 29.7\% have Setswana as the second-best prediction (mean competing language score 0.756, mean Setswana residual score 0.134).
The competing language is predominantly Northern Sotho/Sepedi (\texttt{nso}), which shares significant lexical overlap with Setswana.

\paragraph{Anonymization}
All usernames, mentions, URLs, and other PII information were replaced with placeholder tokens before annotation to protect user privacy.

\subsection{Annotation}
\label{sec:annotation}

\paragraph{Label schema}
Each tweet was assigned one of five labels:
\textbf{Positive}, \textbf{Negative}, \textbf{Neutral}, \textbf{Mixed} (tweet expresses both positive and negative sentiment), or \textbf{Indeterminate} (tweet is unintelligible, in an unknown language, or cannot be confidently categorized).

\paragraph{Annotators.}
Three undergraduate students of African languages annotated the dataset.
All three are native speakers of Setswana.
Before beginning the proper annotation task, the annotators completed a two-round training phase on 65 tweets, with adjudication and discussion of disagreements between rounds. The annotators are referred to as Ann.~A, Ann.~B, and Ann.~C throughout.

\paragraph{Annotation tool and batching}
Annotations were collected through the LightTag annotation tool \cite{perry-2021-lighttag}, now part of the Primer.ai NLP platform. The platform recorded the precise UTC timestamp of each label submission and other metadata such as annotator identity, per-example view logs (recording which annotators were assigned each item), a review and adjudication slot, and a free-text comment field (unused in our campaign).
Production annotation was divided into seven batches of 500 tweets each (Batches~1--7), completed sequentially.
Each annotator labeled all tweets in a batch independently and asynchronously, without access to the labels assigned by the other two annotators.

\subsection{Dataset Statistics}
\label{sec:stats}

\begin{figure}[t]
  \centering
  \includegraphics[width=\linewidth]{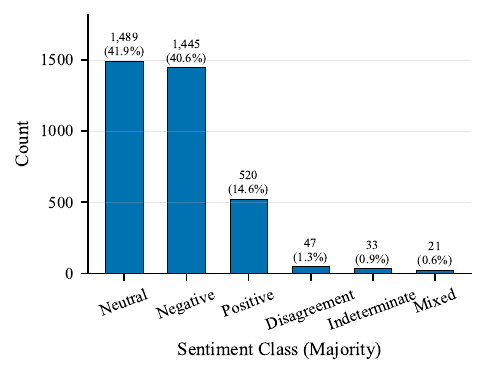}
  \caption{Sentiment label distribution after adjudication.}
  \label{fig:sentiment_dist}\vspace{-5.5mm}
\end{figure}

The full corpus contains 3,565 annotated tweets.
For tweets with unanimous three-way agreement, the label is taken directly; for 2-of-3 majority agreement, the majority label is retained; tweets with full three-way label conflict are excluded from downstream classification.
The final released dataset contains 520 Positive, 1,445 Negative, and 1,489 Neutral tweets. Mixed, Indeterminate, and fully-conflicted labels together account for less than 3\% of all examples and are not included in the classification split.
\Cref{fig:sentiment_dist} shows the overall label distribution.

\section{Annotation Quality Analysis}
\label{sec:quality}

We conduct six targeted analyzes using the full annotation logs, including per-annotator timestamps for every label submission.
All analyzes are restricted to tweets for which all three annotators provided a label (3,555 of 3,565 tweets; 10 edge-case tweets at batch boundaries are excluded).

\subsection{RQ1: $\kappa$ Over Batches}
\label{sec:exp1}

\begin{figure}[t]
  \centering
  \includegraphics[width=\linewidth]{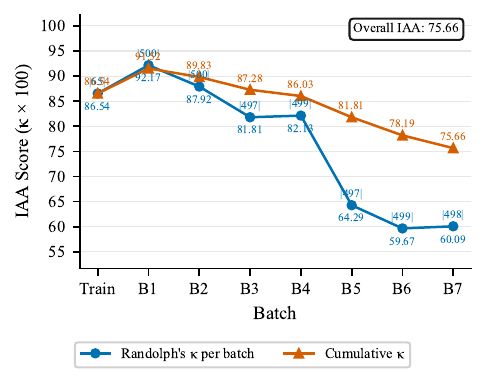}
  \caption{Per-batch Randolph's $\kappa$ (blue) and cumulative $\kappa$ (orange) across the training batch and Batches~1--7. Batch sizes are shown as $|n|$ above each point. $\kappa$ declines monotonically from Batch~2 onward; the steepest drop is between Batch~4 and Batch~5.}
  \label{fig:iaa_over_time}\vspace{-5.5mm}
\end{figure}

\Cref{fig:iaa_over_time} shows both the per-batch $\kappa$ and the cumulative $\kappa$ as each batch is added. Per-batch $\kappa$ starts high ($\kappa = 92.17$ for Batch~1) and declines to $\kappa = 60.09$ by Batch~7, a fall of more than 32 points. The sharpest single drop occurs between Batch~4 ($\kappa = 82.13$) and Batch~5 ($\kappa = 64.29$), a 17.8-point decline in a single batch. The cumulative $\kappa$ ends at 75.66, which is technically "excellent" by Randolph's thresholds but masks the severity of the deterioration in the final three batches. We investigate this decline from five angles in the sections that follow.

\subsection{RQ2: Label Confusion}
\label{sec:exp2}

\begin{figure*}[t]
  \centering
  \begin{subfigure}{0.32\textwidth}
    \includegraphics[width=\linewidth]{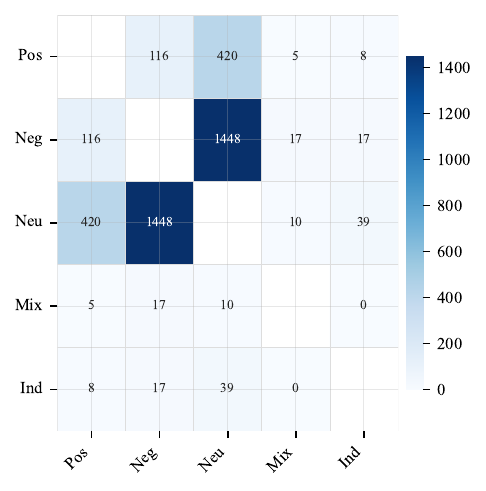}
    \caption{All batches}
  \end{subfigure}\hfill
  \begin{subfigure}{0.32\textwidth}
    \includegraphics[width=\linewidth]{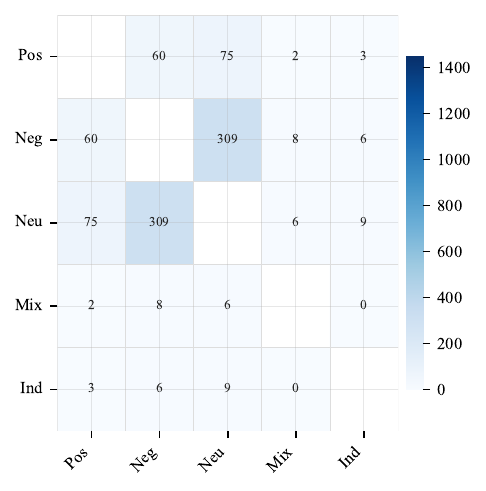}
    \caption{Early (Train--B3)}
  \end{subfigure}\hfill
  \begin{subfigure}{0.32\textwidth}
    \includegraphics[width=\linewidth]{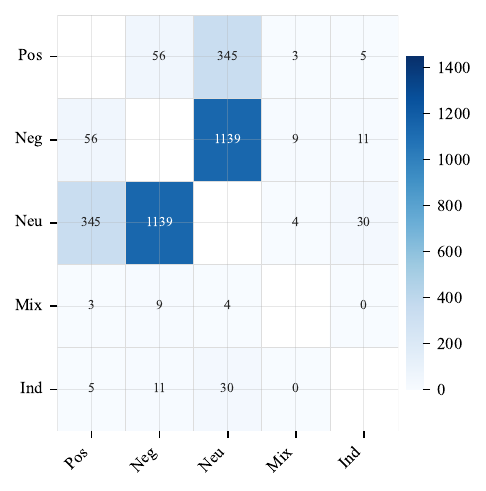}
    \caption{Late (B4--B7)}
  \end{subfigure}
  \caption{Pairwise annotator label confusion heatmaps. Each cell $(i, j)$
    counts annotator pairs where one chose label $i$ and the other
    chose label $j$; diagonal cells are masked.
    All three heatmaps share the same colour scale.
    The Negative/Neutral cell (1,448 cases) dominates across all
  periods and intensifies in the late batches.}
  \label{fig:confusion}\vspace{-5.5mm}
\end{figure*}

To understand which distinctions annotators find hardest, we construct a pairwise confusion matrix, counting, for each annotator pair on each tweet, how often two annotators assigned different labels.
\Cref{fig:confusion} shows this matrix aggregated over all batches and split into early (Training--Batch~3) and late (Batch~4--7) periods.

The dominant confusion is overwhelmingly \textbf{Negative vs.\ Neutral} (1,448 pairwise cases), an order of magnitude larger than the next most frequent pair, Positive vs.\ Neutral (420 cases).
This pattern is linguistically interpretable: Setswana political and social commentary frequently uses indirect or ironic phrasing where the boundary between a statement of fact (Neutral) and a negatively-valenced judgement (Negative) is genuinely ambiguous. This ambiguity is a property of the language and content domain, not solely of annotator quality. This motivates retaining these instances rather than discarding them.

Comparing the early and late heatmaps, the Negative/Neutral cell darkens in later batches while confusions involving Mixed and Indeterminate remain negligible throughout. This is consistent with annotators applying less careful distinctions as the campaign progresses, rather than simply encountering harder tweets.

\subsection{RQ3: Autopilot Behavior}
\label{sec:exp3}

A common symptom of annotation fatigue is \textbf{\emph{autopilot}} behavior, where the annotator repeatedly applies the same label without deliberation, producing long uninterrupted runs of identical labels in timestamp order.
We calculated this as the "\textbf{\texttt{mean same-label run length}}" per annotator per batch, an uninterrupted sequence of identical consecutive labels in annotation submission order.

\begin{figure}[t]
  \centering
  \includegraphics[width=\linewidth]{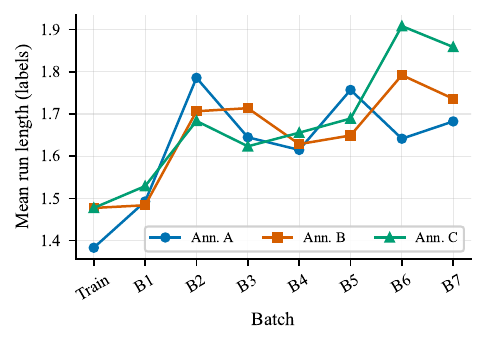}
  \caption{Mean same-label run length per annotator per batch. All three
    annotators trend upward; the increase is most pronounced and consistent
    for Ann.~B and Ann.~C, particularly from Batch~5 onward.}
  \label{fig:run_length}\vspace{-4.5mm}
\end{figure}

\Cref{fig:run_length} shows that mean run length increases for all three annotators across batches, starting near 1.4--1.5 labels per run in the training batch and rising to 1.7--1.9 by Batch~6 to 7. Mean run length increases consistently across batches for Ann.~B and Ann.~C, reaching its highest values in Batches~6 and 7, while Ann.~A shows a weaker and less consistent trend.
The upward pattern for two of the three annotators indicates that the labelling pattern, not just the labels themselves, changes over the annotation process.

Longer runs of length $\geq 5$ provide a conservative indicator of non-deliberate labelling (\Cref{fig:long_runs} in Appendix~\ref{sec:appendix_long_runs}). The mean count of such runs per annotator grows from Batch~1 to Batch~7, with the increase concentrated in Ann.~B and Ann.~C from Batch~5 onward. These are the same batches where $\kappa$ falls most steeply.

\subsection{RQ4: Annotation Speed}
\label{sec:exp4}

\begin{figure}[t]
  \centering
  \includegraphics[width=\linewidth]{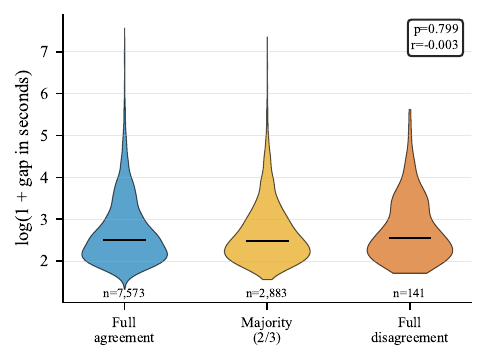}
  \caption{Distribution of log-transformed inter-annotation gaps (seconds) by tweet-level agreement category. Medians (horizontal bars) are virtually identical across all three groups, indicating no relationship between annotation speed and agreement.}
  \label{fig:speed_violin}\vspace{-4.5mm}
\end{figure}

We compute the inter-annotation gap for each annotator as the elapsed time in seconds between consecutive label submissions in timestamp order. \Cref{fig:speed_violin} compares the gap distribution across three agreement categories at the tweet level. The gap distributions of fully-agreed, majority-agreed, and fully-disagreed tweets are nearly identical, with near-overlapping medians and shapes (\Cref{fig:speed_violin}).

\begin{figure}[t]
  \centering
  \includegraphics[width=\linewidth]{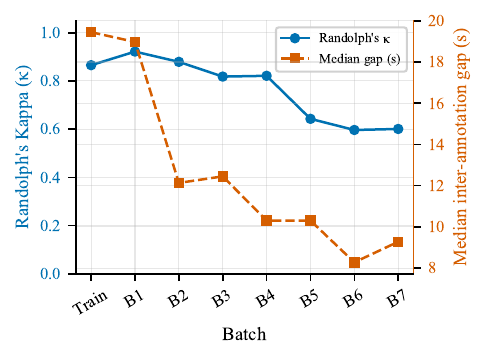}
  \caption{Randolph's $\kappa$ (left axis, blue) and median inter-annotation gap in seconds (right axis, orange dashed) per batch. The two series decouple: speed drops sharply at Batch~2 while $\kappa$ remains stable; $\kappa$ falls steeply at Batch~5 while speed has already plateaued.}
  \label{fig:speed_kappa}\vspace{-3.5mm}
\end{figure}

At the batch level (\Cref{fig:speed_kappa}), median annotation speed does increase over the annotation process. Annotators label tweets progressively faster, with median gap falling from approximately 20s in the training batch to under 9s by Batch~6.
Critically, this acceleration is decoupled from $\kappa$: the largest speed-up occurs between Batch~1 and Batch~2, while $\kappa$ remains stable in that period; conversely, the sharpest $\kappa$ drop (Batch~4$\to$5) occurs after speed has already plateaued. Annotation speed and annotation quality therefore evolve independently.

\subsection{RQ5: Temporal Simultaneity}
\label{sec:exp5}

\begin{figure}[t]
  \centering
  \includegraphics[width=\linewidth]{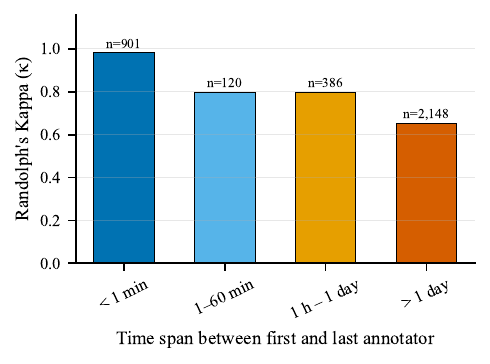}
  \caption{Randolph's $\kappa$ by time-span bin (elapsed time between the first and last annotator's timestamp for the same tweet). Synchronous labelling ($< 1$~min, $n = 901$) yields near-perfect $\kappa = 0.98$; asynchronous labelling ($> 1$~day, $n = 2{,}148$) yields $\kappa = 0.65$.}
  \label{fig:simultaneity_bins}
\end{figure}

Because annotators worked asynchronously, the same tweet might be reviewed by one annotator on one day and another days or weeks later. We define the \textbf{\emph{time span}} for a tweet as the elapsed time between the first and last annotator's timestamp. \Cref{fig:simultaneity_bins} shows the $\kappa$ values after grouping tweets into four bins: $< 1$~min, $1$--$60$~min, $1$~hr--$1$~day, and $> 1$~day. The gradient is very visible:

\begin{itemize}[noitemsep]
  \item $< 1$~min ($n = 901$): $\kappa = 0.98$
  \item $1$--$60$~min ($n = 120$): $\kappa = 0.80$
  \item $1$~hr--$1$~day ($n = 386$): $\kappa = 0.80$
  \item $> 1$~day ($n = 2{,}148$): $\kappa = 0.65$
\end{itemize}

\begin{figure}[t]
  \centering
  \includegraphics[width=\linewidth]{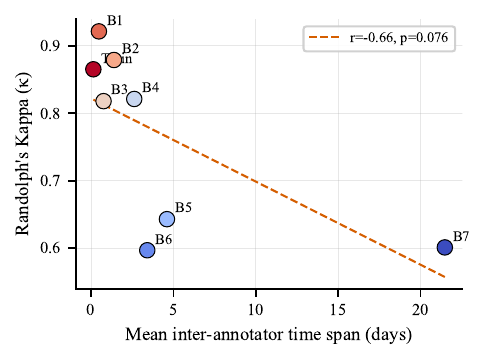}
  \caption{Batch-level mean inter-annotator time span (days) vs. Randolph's $\kappa$. The trend is consistent: batches with longer mean time spans have lower $\kappa$; Batch~7 is a notable outlier at over 20 days mean span.}
  \label{fig:simultaneity_scatter}\vspace{-3.5mm}
\end{figure}

Tweets labeled within a single minute achieve near-ceiling agreement; tweets labeled more than a day apart lose 33 points of $\kappa$ relative to the synchronous group. The majority of tweets in the dataset (60.3\%) fall in the $> 1$~day bin, which means the annotation's asynchronous structure structurally suppresses the overall $\kappa$. At the batch level (\Cref{fig:simultaneity_scatter}), there is a strong negative association between mean inter-annotator time span and $\kappa$. As visible in the scatter, later batches were annotated with substantially greater time spread: Batch~7 has a mean span of over 20~days, compared to less than one day for Batches~1--2, and $\kappa$ tracks this spread closely.

Cross-referencing annotation dates with the 
university academic calendar provides a plausible structural explanation for this pattern. Batches~1--3 (14--28 March) fell during normal lectures; Batch~5 (10--18 April) coincided with the end of Quarter~1 on 17~April, when assignment deadlines and tests are concentrated; and Batch~7 (24 April--24 May) extended into end-of-semester preparation, with two annotators completing their work on 24~May, the day before the university's cooling-off period for non-academic activities began (25~May).
Academic pressure thus provides an external account of why annotators spread their work across progressively more days in later batches, increasing inter-annotator time spans and depressing $\kappa$.

\begin{figure*}[t]
  \centering
  \begin{subfigure}{0.48\textwidth}
    \centering
    \includegraphics[width=0.96\linewidth]{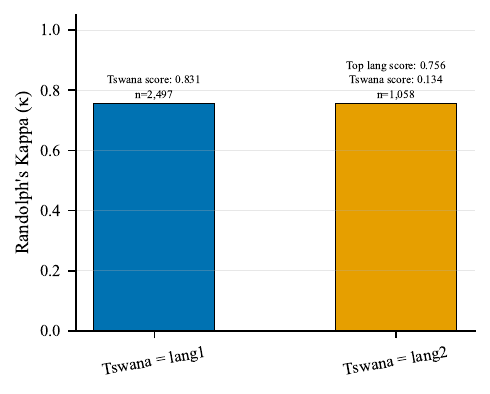}
    \caption{Top-1 (lang1) vs.\ second-best (lang2) LID predictions.}
    \label{fig:lid_kappa_overall}\vspace{-3.5mm}
  \end{subfigure}
  \hfill
  \begin{subfigure}{0.48\textwidth}
    \centering
    \includegraphics[width=\linewidth]{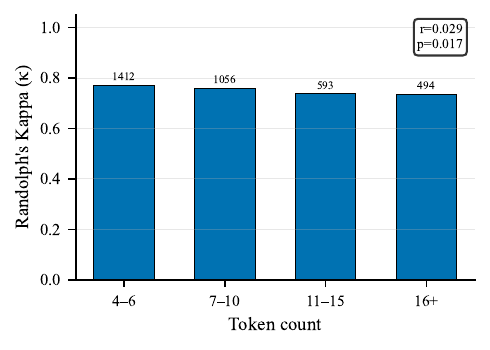}
    \caption{Agreement by token-count bin.}
    \label{fig:token_kappa}\vspace{-3.5mm}
  \end{subfigure}

  \caption{Annotation agreement (Randolph's $\kappa$) is stable across conditions:
    (a) no meaningful difference between top-1 vs.\ second-best LID predictions ($\kappa \approx 0.757$ for both groups);
  (b) minimal variation across token-count bins ($\kappa$ ranges 0.73--0.77 with no consistent trend).}
\end{figure*}

\subsection{RQ6: Tweet Complexity}
\label{sec:exp6}

We examine two tweet-level proxies for annotation difficulty: language identification (LID) rank and token count.

\paragraph{LID rank.}
Tweets where Setswana is the top-1 LID prediction ("lang1", $n = 2{,}497$, mean score $= 0.831$) and those where another language outscores Setswana ("lang2", $n = 1{,}058$, mean competing score $= 0.756$, mean Setswana score $= 0.134$) yield essentially identical $\kappa$: $0.757$ and $0.757$ respectively (\Cref{fig:lid_kappa_overall}).
The agreement rates for the two groups are virtually identical, indicating no association between LID rank and annotator agreement. Per-batch trends are parallel for both groups (\Cref{fig:lid_kappa_batch} in \Cref{sec:appendix_lid}), confirming that LID ambiguity does not differentially affect agreement.

\paragraph{Token count.}
Binning tweets by token count into four ranges (4--6, 7--10, 11--15, 16+) yields $\kappa$ values of approximately 0.77, 0.76, 0.74, and 0.73, respectively, a near-flat profile with no consistent trend (\Cref{fig:token_kappa}). The difference across bins is negligible and tweet length accounts for virtually none of the observed variation in agreement.

Together, the LID and token-count null results establish that $\kappa$ is not primarily driven by tweet difficulty. This shifts explanatory responsibility toward annotator-level factors such as fatigue, drift, and scheduling. These are addressed by RQ3 and RQ5.

\section{Discussion}
\label{sec:discussion}


\subsection{Disagreement Is Not the Enemy}

Before discussing individual findings, we emphasize a framing point. A growing body of work demonstrates that annotator disagreement encodes genuine ambiguity, diversity in perspectives, and domain-specific nuance that majority-vote aggregation actively discards \cite{plank-2022-problem,fleisig-etal-2024-perspectivist,xu-jurgens-2026-beyond}. We take this seriously. The dominant confusion in our dataset, between Negative and Neutral (1,448 pairwise cases, RQ2), reflects a genuine property of Setswana political discourse, where indirect evaluative language straddles the boundary between factual statement and sentiment expression. This disagreement is linguistically meaningful, and we do not treat it as an error to be eliminated.

What we \emph{do} target are two qualitatively different patterns that emerge in later batches: autopilot labelling (RQ3) and scheduling-driven temporal asynchrony (RQ5). These are process artefacts that inflate disagreement beyond whatever the data itself warrants, and they are detectable and correctable without discarding the inherent ambiguity that makes the task interesting. The key empirical observation is that the Negative/Neutral confusion is \emph{stable} across batches (it was always the dominant pair), whereas the $\kappa$ decline in Batches~5--7 is driven by a broad-based increase in confusion across \emph{all} label pairs, consistent with process degradation rather than growing data ambiguity.

\begin{table}[t]
  \centering
  \small
  \resizebox{\columnwidth}{!}{
    \begin{tabular}{p{0.4cm}lp{4cm}}
      \toprule
      RQ & Factor & Finding \\
      \midrule
      1 & Batch trend & $\kappa$ declines 92$\to$60; sharpest at B4$\to$B5 \\
      2 & Label confusion & Neg/Neu dominates (1,448 cases); intensifies in late batches \\
      3 & Autopilot & Run length increases significantly for Ann.~B \& C \\
      4 & Speed & Not a predictor; gap distributions are virtually identical across groups \\
      5 & Simultaneity & Strongest predictor: $\Delta\kappa = 0.33$ from ${<}$1~min to ${>}$1~day \\
      6 & LID / tokens & No meaningful association; $\kappa$ stable across both variables \\
      \bottomrule
    \end{tabular}
  }
  \caption{Summary of findings across the six research questions.}
  \label{tab:rq_summary}\vspace{-3.5mm}
\end{table}

\subsection{Interpretation}

The temporal simultaneity finding (RQ5) is both the largest in magnitude and the most actionable. When all three annotators label a tweet within minutes, they are likely in a shared annotation context, with similar energy levels, similar interpretation of boundary cases, and no intervening exposure to other sessions that might subtly shift their decision criteria. The 33-point $\kappa$ gap between synchronous ($< 1$~min) and asynchronous ($> 1$~day) tweets is far larger than any other factor we examined, and the batch-level pattern confirms the same gradient at a coarser granularity.

The run-length results (RQ3) provide a complementary, behaviorally grounded account. By Batches~5--7, Ann.~B and Ann.~C show statistically significant increases in mean run length and in the count of long streaks ($\geq 5$ labels), suggesting that the decision process has shifted from deliberate per-tweet judgment to more automatic pattern-matching. The sharpest $\kappa$ decline (Batch~4$\to$5) coincides with the period where run-length growth is most pronounced. This points to a common underlying mechanism: reduced annotator engagement, not harder tweets. 

The null result for annotation speed (RQ4) corrects a common intuition. The widely-held belief that "faster $\Rightarrow$ worse" is not supported at the tweet level: gap distributions for agreed and conflicted tweets are virtually identical. The aggregate annotation-level co-occurrence of faster labelling and lower $\kappa$ in later batches therefore reflects a shared cause (reduced engagement) rather than a direct causal chain. 

The null results for LID ambiguity and token count (RQ6) are practically important. They establish that filtering tweets by linguistic complexity would not meaningfully improve $\kappa$, and that the process-driven quality issues are annotator-level rather than content-level. This means the same tweets, re-annotated under better scheduling conditions, would likely achieve substantially higher $\kappa$ because the disagreement in the data is not the problem.

To improve annotation quality in future campaigns, we recommend implementing interventions based on our findings, as presented in Appendix \ref{app:recommnd}.





\section{Classification Benchmark Results}
\label{sec:experiments}

Given the observed annotation quality patterns, we conduct a preliminary evaluation of the dataset's utility for sentiment classification. This serves as a benchmark for future work. The detailed experimental setup is presented in the Appendix \ref{app:setup}.


\begin{table}[t]
  \centering
  \resizebox{\columnwidth}{!}{
    \begin{tabular}{llcc}
      \toprule
      Model & Setting & Macro-F1 & Accuracy \\
      \midrule
      mBERT           & pre-trained (probe) & 20.6 & 24.2 \\
      AfriBERTa       & pre-trained (probe) &  8.7 & 15.0 \\
      AfroXLMR-base   & pre-trained (probe) & 10.6 & 16.1 \\
      \midrule
      mBERT           & fine-tuned & 50.6\small{$\pm$0.9} & 55.5 \\
      AfriBERTa       & fine-tuned & 49.0\small{$\pm$2.1} & 53.8 \\
      AfroXLMR-base   & fine-tuned & \textbf{53.6}\small{$\pm$1.2} & 56.7 \\
      \midrule
      GPT-5           & zero-shot  & 56.8 & 57.6 \\
      Gemini & zero-shot & \textbf{57.1} & 57.4 \\
      \midrule
      GPT-5           & few-shot   & \textbf{62.2} & 62.5 \\
      Gemini & few-shot & 57.2 & 57.4 \\
      \bottomrule
    \end{tabular}
  }
  \caption{Three-class sentiment classification results on our test set (macro-F1 and accuracy, \%). Pre-trained models are evaluated via MLM probing with no task-specific data. Fine-tuned models report mean $\pm$ std over 5 random seeds. Bold indicates best within each group.}
  \label{tab:classification}
\end{table}

\paragraph{Results} \Cref{tab:classification} reveals three key findings. First, pre-trained encoders without task-specific data perform near or below random (8.7--20.6 macro-F1), confirming that multilingual pre-training alone is insufficient for Setswana sentiment prediction, directly motivating the dataset. Second, fine-tuning on our data yields large, consistent gains across all three encoders (+29--43 macro-F1 points), with AfroXLMR-base achieving the best fine-tuned result (53.6), consistent with its broader African-language pre-training \cite{alabi-etal-2022-adapting}. Third, GPT-5 in few-shot mode outperforms all fine-tuned encoders (62.2), suggesting that proprietary LLMs can leverage limited in-context examples effectively even for low-resource languages, though at significantly higher inference cost. Gemini 3, by contrast, shows little benefit from few-shot prompting (57.1 zero-shot vs. 57.2 few-shot), suggesting that in-context examples do not consistently help across frontier models for this low-resource language.
\section{Conclusion}
\label{sec:conclusion}

We have presented a Setswana Twitter sentiment dataset and a systematic, timestamp-driven analysis of annotation quality across an eight-batch annotation process. The dominant predictor of inter-annotator agreement is \textbf{\emph{temporal simultaneity}}: tweets labeled within one minute achieve $\kappa = 0.98$; those labeled more than a day apart reach only $\kappa = 0.65$. Run-length drift in two annotators provides a complementary behavioral account of quality decline.

For low-resource NLP annotation tasks that rely on small pools working independently, precisely the conditions that maximize inter-annotator time spans, batch-scoped annotation windows, and run-length monitoring are low-cost interventions that could substantially improve dataset quality. We release the full dataset, timestamps, and analysis code to support reproducible quality auditing in future African-language NLP.

\section*{Limitations}
\label{sec:limitations}

Several limitations of this work should be noted.
\textbf{Language scope:} The dataset covers Setswana as used on South African Twitter and may not generalize to other dialects or registers of the language.
\textbf{Annotator pool:} Three annotators from a single institution participated in the campaign; a more diverse pool might yield different agreement patterns.
\textbf{Label schema:} The four-class schema (Positive, Negative, Neutral, Mixed) collapses nuanced sentiment; binary or fine-grained schemes could be explored in future work.
\textbf{Platform bias:} Twitter data over-represents urban, younger, and digitally active speakers, which may not reflect the broader Setswana-speaking population.
\textbf{Temporal coverage:} Tweets were collected during a specific period; sentiment distributions may shift with news cycles or social events.
\textbf{LLM evaluation:} LLM results reflect a single evaluation run with fixed prompts; different prompt formulations may yield different results. The gap between GPT-5 and Gemini 3 few-shot performance may also reflect differences in instruction-following rather than underlying language competence.

\section*{Ethics Statement}
\label{sec:ethics}

\textbf{Data source:} All tweets were collected via the official Twitter/X Academic API under its terms of service for academic research. We do not release tweet IDs alongside raw text; instead, we release cleaned text only, in line with common practice for low-resource NLP datasets where data scarcity makes ID-only release impractical for reproducibility.
\textbf{Annotator welfare:} Annotators were 
university students who participated voluntarily and were compensated at the official university hourly rate. No sensitive personal information was collected beyond the annotation labels.
\textbf{Ethical clearance:} This project received ethical clearance from the 
University Research Ethics Committee before data collection and annotation.
\textbf{Potential harms:} Sentiment classifiers trained on this data could be misused for surveillance or opinion mining of Setswana-speaking individuals. We release the dataset for research purposes and encourage users to consider these risks before deployment.
\textbf{Broader impact:} Setswana is an under-resourced language; we hope this dataset lowers the barrier to NLP research for the language and contributes to more equitable representation in multilingual AI.

\section*{Dataset License}
The dataset is released under a controlled-access license, NOODL,\footurl{https://licensingafricandatasets.com/nwulite-obodo-license} to promote equitable and responsible use. While permissive licenses such as CC BY maximise accessibility, they do not ensure reciprocal benefit to data contributors, particularly in low-resource language contexts where annotation labour is locally situated but downstream value is often captured by external actors. Our licensing approach aims to balance research accessibility with fairness: non-commercial research use is supported, while commercial use requires engagement with the dataset creators. This reflects a commitment to reciprocity and to ensuring that the benefits of Setswana language resources are more equitably distributed.

\bibliography{anthology,custom}

\appendix

\section{Annotation Guidelines Summary}
\label{sec:appendix_guidelines}

Annotators were provided with a written guideline document covering:
(i) definitions of each sentiment category with six worked examples per category;
(ii) decision rules for the Negative/Neutral boundary (factual statements about negative events are Neutral; explicit evaluative language is Negative);
(iii) rules for code-switched content (label according to the dominant-language sentiment); and
(iv) examples of Indeterminate tweets including non-Setswana content and unintelligible text.

\section{Long Run Counts (RQ3)}
\label{sec:appendix_long_runs}

\begin{figure}[H]
  \centering
  \includegraphics[width=\linewidth]{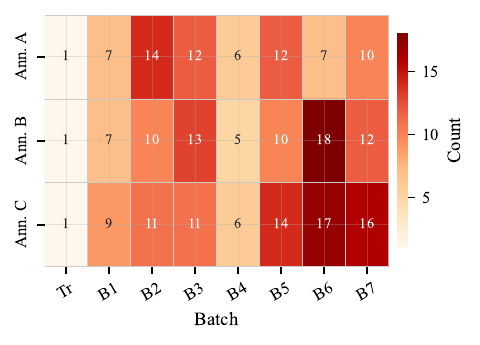}
  \caption{Count of runs of length $\geq 5$ per annotator per batch. Darker shading indicates more long streaks of the same label. The increase is concentrated in Ann.~B and Ann.~C from Batch~5 onward, the same batches where $\kappa$ falls most steeply.}
  \label{fig:long_runs}
\end{figure}

\section{LID Null Result (RQ6)}
\label{sec:appendix_lid}

\begin{figure}[H]
  \centering
  \includegraphics[width=\linewidth]{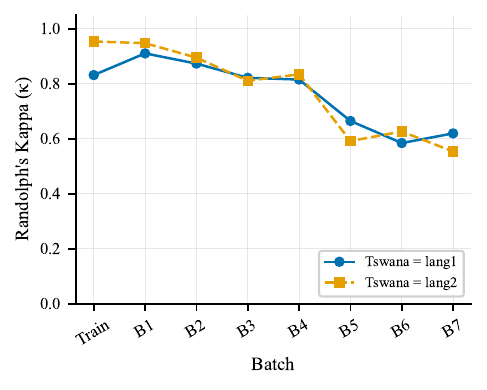}
  \caption{Per-batch $\kappa$ for the lang1 and lang2 LID groups. The two lines track each other closely across all batches, confirming that LID ambiguity does not differentially drive quality decline.}
  \label{fig:lid_kappa_batch}
\end{figure}

\section{Cumulative Annotation Progress per Batch}
\label{sec:appendix_rest}

\Cref{fig:accel_2} plots cumulative annotations against wall-clock time, a distance-time graph where slope encodes speed and flat segments are rest periods.
Per-session speed is labeled on each active segment; red arrows mark extreme idle gaps (above the 75th percentile for that panel, minimum 4~h).
Later batches show increasing fragmentation into more and longer rest periods, consistent with annotators spreading work across more sessions and the consequent depression of $\kappa$ documented in RQ5.

\begin{figure*}[!ht]
  \centering
  \begin{subfigure}{\textwidth}
    \includegraphics[width=\textwidth]{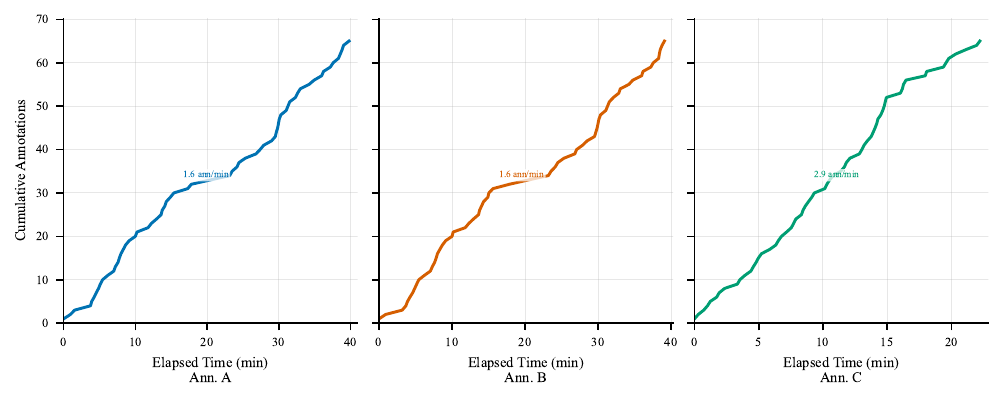}
    \subcaption{Training batch}
  \end{subfigure}
  \begin{subfigure}{\textwidth}
    \includegraphics[width=\textwidth]{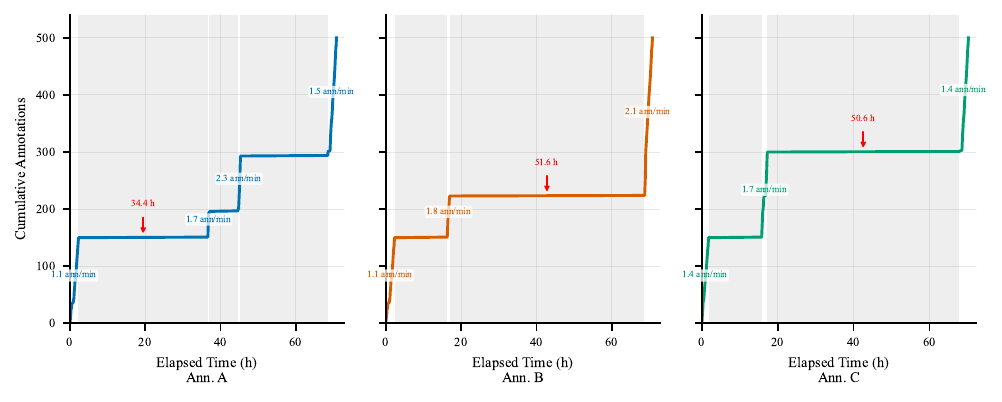}
    \subcaption{Batch 1}
  \end{subfigure}
  \begin{subfigure}{\textwidth}
    \includegraphics[width=\textwidth]{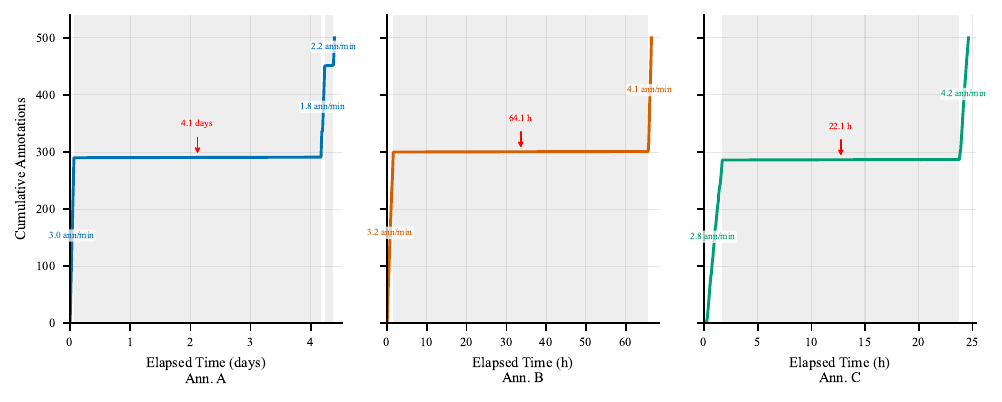}
    \subcaption{Batch 2}
  \end{subfigure}
  \label{fig:accel_1}
\end{figure*}

\begin{figure*}[!ht]\ContinuedFloat
  \centering
  \begin{subfigure}{\textwidth}
    \includegraphics[width=\textwidth]{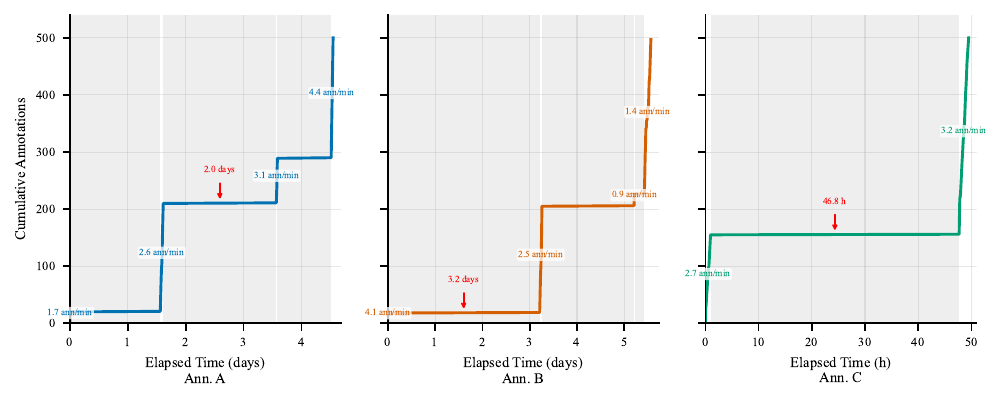}
    \subcaption{Batch 3}
  \end{subfigure}
  \begin{subfigure}{\textwidth}
    \includegraphics[width=\textwidth]{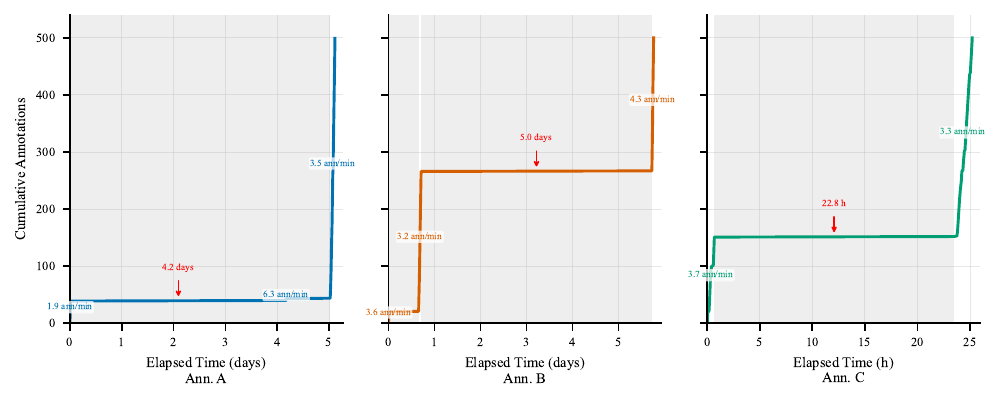}
    \subcaption{Batch 4}
  \end{subfigure}
  \begin{subfigure}{\textwidth}
    \includegraphics[width=\textwidth]{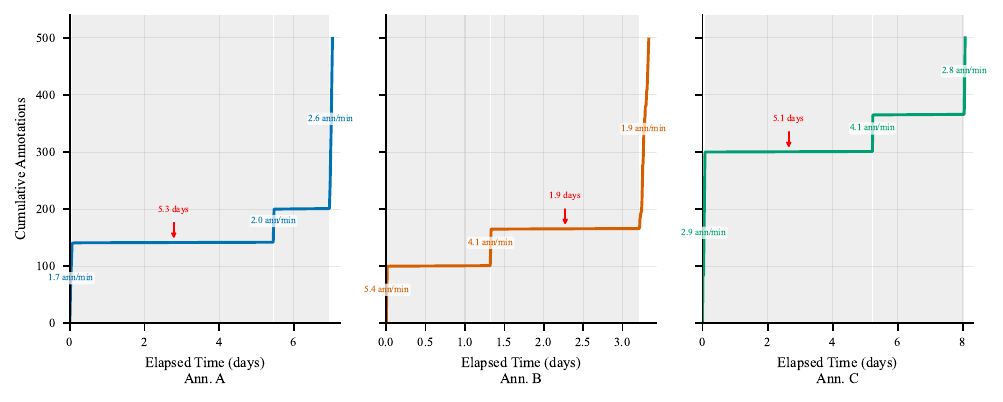}
    \subcaption{Batch 5}
  \end{subfigure}
  \label{fig:accel_1}
\end{figure*}

\begin{figure*}[!ht]\ContinuedFloat
  \begin{subfigure}{\textwidth}
    \includegraphics[width=\textwidth]{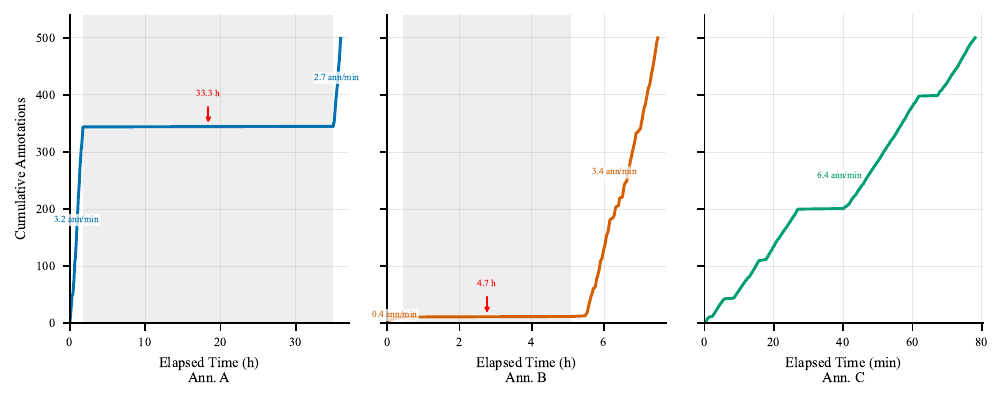}
    \subcaption{Batch 6}
  \end{subfigure}
  \begin{subfigure}{\textwidth}
    \includegraphics[width=\textwidth]{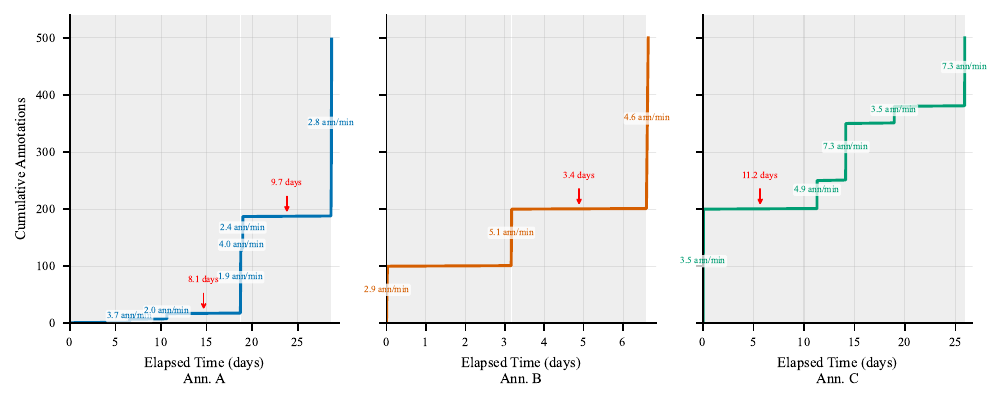}
    \subcaption{Batch 7}
  \end{subfigure}
  \caption[]{Cumulative annotation progress per annotator per batch.
    X-axis: elapsed time (min / h / days); y-axis: cumulative tweets annotated.
    Grey shading = rest period ($\geq 1$~h idle); red arrows = extreme idle gaps
  (duration at arrow base); coloured labels = per-session speed (ann/min).}
  \label{fig:accel_2}
\end{figure*}

\section{Experiment Setup}\label{app:setup}

We evaluate the dataset on three-class sentiment classification (Positive, Negative, Neutral) using the majority-vote labels from \Cref{sec:stats}, with an 80/10/10 train/dev/test split stratified by label.

Three multilingual pre-trained language models are evaluated: mBERT \cite{devlin-etal-2019-bert} as a general multilingual baseline; AfriBERTa \cite{ogueji-etal-2021-small}, and AfroXLMR-base \cite{alabi-etal-2022-adapting}, both adapted to African languages. Each model is first zero-shot evaluated via masked language modeling (MLM): the sentiment word in a fixed template is replaced with \texttt{[MASK]} and the highest-probability candidate determines the predicted label.

Fine-tuning follows the procedure of \citet{shode-etal-2022-yosm} adapted to our three-class setup. We use AdamW with a learning rate of $5 \times 10^{-5}$, a linear schedule with 6\% warmup steps, batch size 32, and maximum sequence length 128. Models are trained for up to 20 epochs; training is stopped early if dev macro-F1 does not improve for 3 consecutive epochs, and the best checkpoint by dev macro-F1 is used for final test evaluation. Gradients are clipped at a norm of 1.0. Each model is trained with five random seeds; we report macro-F1 and accuracy averaged over seeds. We additionally evaluate GPT-5 \cite{openai-gpt5-2025} and Gemini 3 Flash Preview \cite{gemini3} in zero-shot and six-shot (2 examples per class) settings.



\section{Recommendations}
\label{app:recommnd}
To improve annotation quality in future campaigns, we recommend the following interventions based on our findings:

\begin{enumerate}[noitemsep]
  \item \textbf{Maximize temporal simultaneity.} Requiring all annotators to complete each batch within a short fixed window (e.g., 48 hours) would reduce mean inter-annotator time spans and is the single highest-leverage intervention our results support. Where feasible, shared online annotation sessions, where all annotators label the same tweets in real time, could bring $\kappa$ closer to the $\geq 0.98$ we observe for the $< 1$~min group.

  \item \textbf{Monitor run lengths between batches.} Mean run length is detectable before a batch is finalized and requires only the annotation log. A simple threshold alert (e.g., mean run length ${>}1.8$) could trigger a calibration discussion or rest break before quality degrades further.

  \item \textbf{Include periodic calibration items.} Re-annotating a small fixed set from the training batch at the start of each new proper annotation batch provides an early-warning signal of label drift and allows re-calibration before it compounds.

  \item \textbf{Report per-batch $\kappa$ as standard practice.} Aggregate $\kappa$ over a multi-batch campaign can be misleadingly high even when the final batches are severely degraded. We recommend per-batch $\kappa$ as a mandatory element of annotation quality reporting in African-language NLP dataset papers.
\end{enumerate}
\end{document}